\pdfoutput=1

\documentclass[11pt]{article}

\usepackage{acl}

\usepackage{times}
\usepackage{latexsym}
\usepackage{amsmath,amsfonts,amssymb}
\usepackage{times}
\usepackage{latexsym}
\usepackage{booktabs}
\usepackage{multirow}
\usepackage{array}
\usepackage{tikz}
\usetikzlibrary{bayesnet}
\usetikzlibrary{arrows}
\usepackage{float}
\usepackage{graphicx}
\usepackage{subcaption}
\usepackage{svg}
\usepackage[normalem]{ulem}
\usepackage{bbm}
\usepackage{xspace}
\usepackage{setspace}
\usepackage{boxedminipage}

\usepackage[T1]{fontenc}

\usepackage[utf8]{inputenc}

\usepackage{microtype}

\newcommand{\cM}{\mathcal{M}}

\newcommand{\figref}[1]{Figure~\ref{#1}}
\newcommand{\secref}[1]{Section~\ref{#1}}
\newcommand{\tabref}[1]{Table~\ref{#1}}
\newcommand{\utt}[1]{{\fontfamily{cmss}\selectfont{#1}}}
\newcommand{\taskname}{\textsc{FlightPref}}

\newcommand{\speakerbase}{\ensuremath S_{\text{base}}}
\newcommand{\listenerbase}{\ensuremath L_{\text{base}}}

\newcommand{\ie}{i.e.,\xspace}
\newcommand{\eg}{e.g.,\xspace}

\usepackage{ulem}

\newif\ifcomments
\commentstrue 
\ifcomments
    \newcommand{\jessy}[1]{{\protect\color{orange}{[JL: #1]}}}
    \newcommand{\dfried}[1]{{\protect\color{blue}{[DF: #1]}}}

    \newcommand{\todo}[1]{{\color{red}{TODO: #1}}}

\else
    \newcommand{\jessy}[1]{}
    \newcommand{\dfried}[1]{}
    \newcommand{\todo}[1]{}
\fi

\title{Inferring Rewards from Language in Context}

\author{
Jessy Lin$^{\diamondsuit}$ \hspace{0.4cm} Daniel Fried$^{\clubsuit}$ \hspace{0.4cm} Dan Klein$^{\diamondsuit}$ \hspace{0.4cm} Anca Dragan$^{\diamondsuit}$ \\
$^{\diamondsuit}$ University of California, Berkeley \\
$^{\clubsuit}$ Carnegie Mellon University\\
\texttt{\{jessy\_lin, klein, anca\}@berkeley.edu, dfried@andrew.cmu.edu}}

\begin{document}
\maketitle
\begin{abstract}
In classic instruction following, language like ``I'd like the JetBlue flight'' maps to actions (\eg selecting that flight).  However, language also conveys information about a user's underlying reward function (\eg a general preference for JetBlue), which can allow a model to carry out desirable actions in new contexts. We present a model that infers rewards from language pragmatically: reasoning about how speakers choose utterances not only to elicit desired actions, but also to reveal information about their preferences.
On a new interactive flight--booking task with natural language, our model more accurately infers rewards and predicts optimal actions in unseen environments, in comparison to past work that first maps language to actions (instruction following) and then maps actions to rewards (inverse reinforcement learning).
\end{abstract}

\section{Introduction}

Language is a natural interface for systems like robots or personal assistants that interact with human users. One way to interpret language in these interactive settings is to train an instruction following agent: a model that learns to map commands like \utt{``go three steps forward to the door''} to a sequence of actions in context (\eg \citealt{Branavan09PG,tellex2011understanding}, \emph{inter alia}). Instructions describe \emph{how} an agent should act in an immediate context, but to build models that can generalize---carrying out a user's goals in new contexts and learning user preferences over repeated interactions---agents should also infer \emph{why} actions are taken. Grounding language to \textit{reward functions} extends the standard instruction following setup in this way, representing the goals and preferences that underlie actions, and allowing agents to autonomously carry out correct actions in new contexts (\eg \citealt{fu2019language}).

\begin{figure}[t!]
    \centering
    \includegraphics[width=\linewidth]{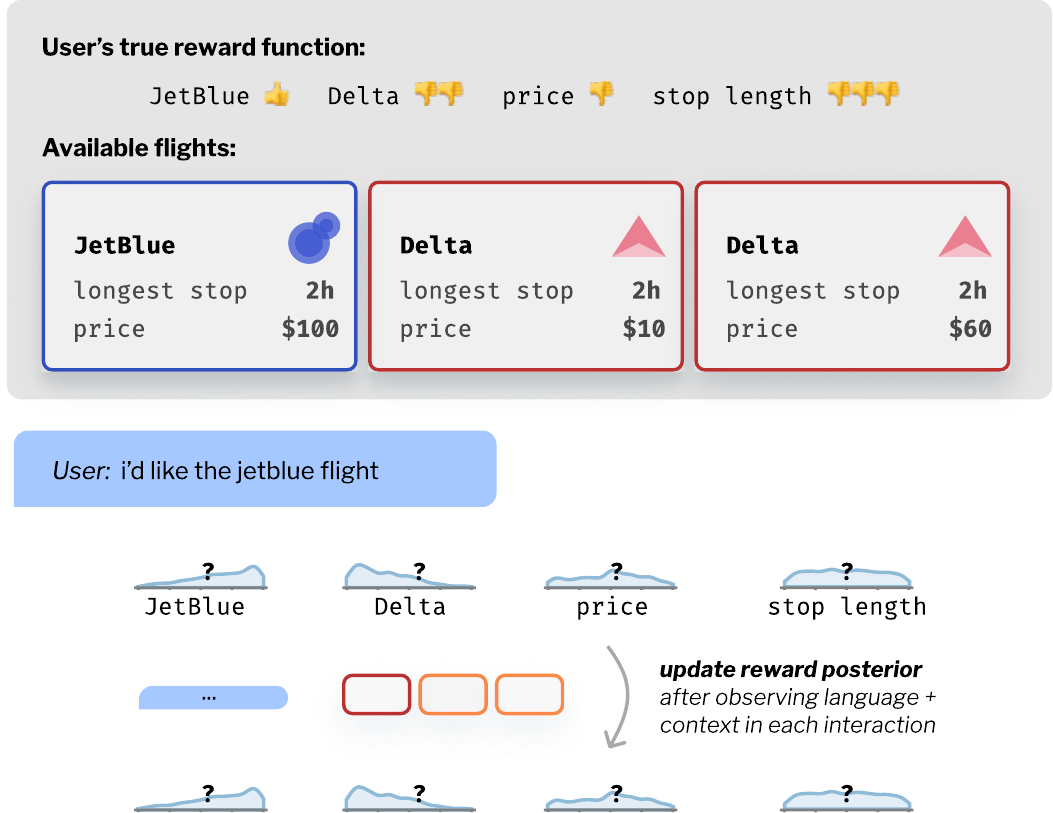}
    \caption{\textbf{When people instruct agents with language like ``I'd like the JetBlue flight,'' both their desired actions and the language itself reveal information about rewards.} From the referenced flight itself, a model would guess that the user may prefer expensive JetBlue flights. Reasoning jointly with language reveals that JetBlue is the more salient preference, and the model should still have uncertainty about whether expensive flights are generally preferred. JetBlue may have been more important than a preference for cheap flights, but the user may still prefer cheap flights, all else equal. Over repeated interactions with the user in new contexts, the model can continually refine its estimates of the user's preferences.}
    \vspace{-1.5em}
    \label{fig:teaser}
\end{figure}
However, when people interact with systems they often primarily aim to achieve specific tasks, rather than literally describing their preferences in full. How do we infer general goals and preferences from utterances in these settings?
Consider a flight booking agent like the one in \figref{fig:teaser}. By inferring the user's reward function (indicating their preference for carrier, price, and other flight features) beyond just selecting the right flight, such a system would be able to autonomously book flights on behalf of the user in other instances. To do so, the system might use the actions the user commands as evidence about what they prefer, recovering rewards from actions using (language-free) techniques like inverse reinforcement learning (IRL; \citealt{ng2000irl}). For example, the system can select a flight the user might like in a new instance by matching features from their past flight bookings.

The key idea of our work is that the \emph{way} that a user refers to their desired actions with language also reveals important information about their reward: the fact that they said ``\utt{the JetBlue flight}'' and not ``\utt{the expensive flight}'' conveys what matters to them. Intuitively, in settings with repeated interactions, utterances are optimized to communicate information that is generalizable---implicitly helping listeners make useful inferences for acting on a longer horizon.
We implement this idea with a pragmatic model of how speakers (humans) generate such language: speakers choose utterances that both elicit reward-maximizing actions in a particular context and faithfully describe the reward. Given an utterance, our model infers that the most likely rewards are the ones that would have made a speaker likely to choose that utterance.

To evaluate our model, we construct and release a dataset for mapping language to rewards, \taskname{}, containing natural language utterances from humans with underlying preferences. Humans interact in a multi-turn flight booking game similar to \figref{fig:teaser}, where we provide a ``user'' player with a reward function representing flight preferences. The goal of the game is for the user to communicate these preferences in natural language to an ``assistant'' player, who is tasked with booking preferred flights for the user. We present this dataset as a challenging benchmark for reward learning from language and interaction.

In our experiments, we show that our model can infer reward functions from natural language, improve reward estimates consistently over repeated interactions, and use inferred rewards to accurately select optimal actions in held-out environments. 
Our full model obtains relative accuracy improvements of 12\% when compared to models that only treat language as descriptions of actions.\footnote{We release our code and dataset at \url{https://github.com/jlin816/rewards-from-language}.}
\section{Related Work}
\paragraph{Instruction following.}
A long line of work on grounded instruction following has developed various methods for producing actions from language, including approaches that use intermediary structured semantic representations \cite{macmahon2006walk,tellex2011understanding,chen2011navigation,matuszek2013learning,artzi2013instructions,she-etal-2014-back,thomason2015learning,wang2016learning,fried2017unified,arumugam2017accurately,suhr2018learning} or map directly to primitive actions \cite{Branavan09PG,andreas-klein-2015-alignment,Mei16Instructions,bisk-etal-2016-natural,misra2017mapping,guu2017bridging,suhr2018situated,anderson2018vision,shridhar2020alfred}.
All of these approaches interpret any given utterance (instruction) solely in the context that elicited the utterance, producing one particular sequence of actions. The method we present extends these approaches, using utterances to infer the rewards that underlie the actions that should be taken across a range of environments: both the context that elicited the utterance, and other unseen environments.

\paragraph{Reward learning.}
The majority of work on reward learning has been in the robotics and reinforcement learning communities and has not incorporated language, rather using techniques such as inverse reinforcement learning (IRL; \citealt{ng2000irl,ratliff2006maximum,ziebart2008maximum,hadfield2017inverse,jeon2020reward}) to infer the rewards that underlie human demonstrations of actions.
Even works that incorporate language into reward learning also take this primarily action-centric approach: either by using datasets pairing utterances with \emph{trajectories} and using (language-free) IRL to then recover reward functions from trajectories \cite{macglashan2015grounding,fu2019language}, or learning an instruction-following model guided by a language-conditioned discriminator \cite{bahdanau2018learning}.
The language in these settings are unambiguous commands, giving a complete description of a goal (\eg ``go to the red door''). In contrast, we are concerned with language used to guide agents in repeated interactions (where language may be a partial or ambiguous mix of instructions and reward descriptions).

\begin{figure*}[ht]
    \centering
    \includegraphics[width=\textwidth]{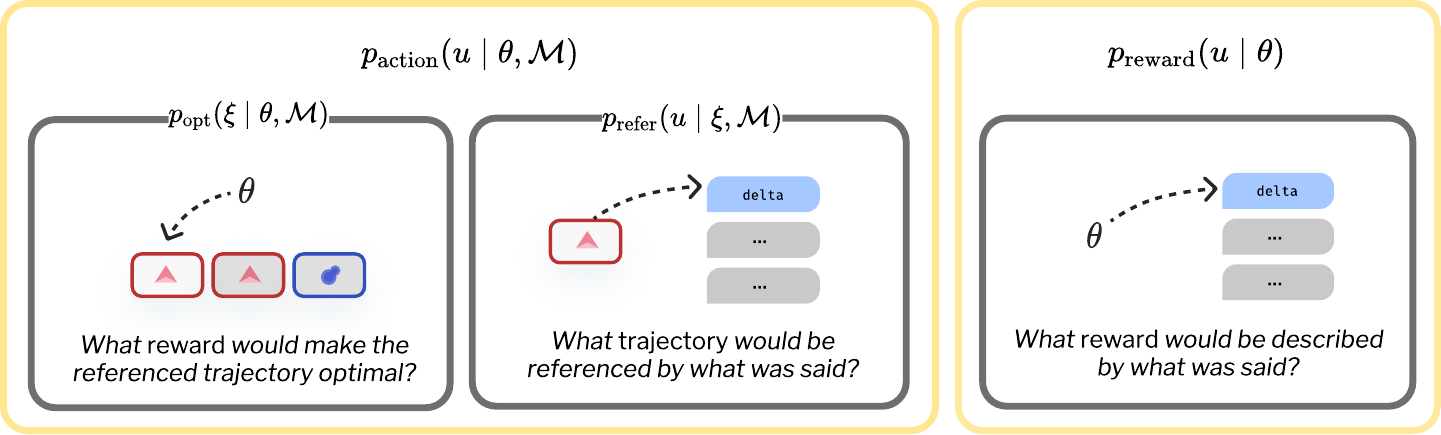}
    \caption{Our model infers rewards by reasoning about how the speaker chose the observed utterance: both to elicit correct actions ($p_{\text{action}}$) and to describe their reward ($p_{\text{reward}}$). We illustrate this on the flight domain, where trajectories are a choice of a single flight.}
    \label{fig:main-model}
    \vspace*{-1em}
\end{figure*}

\paragraph{Pragmatics.}
A long line of work on pragmatics \cite{grice1975logic}, particularly in the Rational Speech Acts (RSA) framework \cite{GoodmanFrank2016-TICS}, has developed computational models for inferring the behavior or belief that a speaker wishes to induce in a listener.
However, the majority of this work has only focused on single-turn interactions, where an utterance conveys an action in a single context, \eg choosing the correct referent in signaling games \cite{Golland10Game,Frank12predictingpragmatic,degen2013cost,monroe2017colors,mcdowell-goodman-2019-learning}, interpreting implicatures \cite{goodman2013knowledge,bergen2016pragmatic}, or generating \cite{fried2017unified,sumers2021extending} or interpreting grounded instructions \cite{fried2018speaker}.
Our work extends this past work by showing that in repeated interactions, listeners can also benefit by reasoning pragmatically about how speakers communicate information about and over longer time horizons.
\section{Reward Inference from Language}
\label{sec:problem-setup}
\paragraph{Problem Formulation.} We parameterize the user's preference as a reward function $r_{\theta}$ with parameters $\theta$. In our flight booking domain from \figref{fig:teaser}, $\theta$ is a weight vector which specifies preferences over flight features (carrier, price, etc.). 
We formalize the general reward inference problem as sequence of Markov decision processes (MDPs) $\cM_1, \ldots, \cM_I$ that share the same reward function $r_{\theta}$.
In each MDP $\cM_i$, the agent receives an utterance $u_i$ from the user and must execute a trajectory $\xi$.
The agent's goal is to infer $\theta$ over the sequence of interactions, which should allow the agent to execute trajectories with high reward in as-yet unseen contexts.

The agent maintains an estimate over $\theta$ over the course of interactions. We introduce a model $p(\theta \mid u, \cM)$ that the agent will use to perform Bayesian updates of a posterior over $\theta$:
\begin{align*}
p(\theta \mid u_{1:i}, \cM_{1:i}) \propto&~p(\theta \mid u_i, \cM_i) \\ 
&\times p(\theta \mid u_{1:i-1}, \cM_{1:i-1})
\end{align*}

In the flight domain, we specialize this formulation to study a one-step MDP (contextual bandit). Trajectories $\xi$ consist of a single action, choosing one of the available flights. Over a series of these rounds where the agent books a flight given the user's utterance $u_i$, the agent must infer the user's flight preferences $\theta$ to book flights from other unseen sets of options, without explicit language instruction from the user.

\subsection{Model}
\label{sec:model}
Our model, summarized in \figref{fig:main-model}, defines a \emph{rational listener}, $L_2$, which predicts a distribution over rewards $\theta$, conditioned on an utterance $u$ and a context $\cM$. (The terminology we use for listeners and speakers follows \citealt{bergen2016pragmatic}.) The rational listener uses Bayesian reasoning about a speaker model, $S_1$, which produces utterances conditioned on a reward function and context:
\[
    p_{L_2}(\theta \mid u, \cM) \propto p_{S_1}(u \mid \theta, \cM) p(\theta \mid \cM)
\]

Key to our model is that the $S_1$ speaker distribution $p_{S_1}(u\mid\theta, \cM)$ defines how speakers produce language that functions both to elicit correct actions and describe their underlying reward:
\begin{multline*}
    p_{S_1}(u \mid \theta, \cM) = 
        \alpha p_{\text{action}}(u \mid \theta, \cM) \\
        + (1-\alpha) p_{\text{reward}}(u \mid \theta),
\end{multline*}
where $\alpha$ controls the speaker's ``nearsightedness''---how much does the speaker care about the listener choosing the correct action in the \emph{current} context, rather than describing the reward in a context-independent way so that the agent can make good choices in \emph{future} contexts?
 
\paragraph{Optimizing for action.} The behavior-optimizing term $p_{\text{action}}$ specifies that the speaker chooses utterances that elicit reward-maximizing behavior from a listener in the current environment:
\begin{align*}
    p_{\text{action}}&(u \mid \theta, \cM) \\ 
    &= \sum_{\xi} p_{\text{refer}}(u \mid \xi, \cM) p_{\text{opt}}(\xi \mid \theta, \cM),
\end{align*}
where the \emph{optimality model} $p_{\text{opt}}(\xi \mid \theta, \cM)$ specifies the probability the speaker refers to trajectory $\xi$ if their true reward is $\theta$.
We can formulate the optimality model with the Boltzmann distribution common in IRL,
where speakers are noisily-rational about which trajectories to refer to: $p_{\text{opt}}(\xi \mid \theta, \cM)\propto \exp(\beta r_{\theta}(\xi;\cM))$, with rationality parameter $\beta$. This term specifies that utterances are more likely to refer to trajectories that have high reward according to the speaker's $\theta$, compared to other trajectories in $\cM$.

Then, for a particular trajectory $\xi$, $p_{\text{refer}}(u \mid \xi,\cM)$ specifies what utterances are likely to refer to that trajectory. In particular, we model that speakers choose utterances that would make a listener execute that trajectory:
\[
p_{\text{refer}}(u \mid \xi,\cM) \propto p_{L_\text{base}}(\xi \mid u,\cM)
\]
using a \emph{base listener} model $\listenerbase$ of the type common in past work on instruction following.
We provide details on $\listenerbase$ in \secref{sec:base-models}.

\paragraph{Optimizing for reward descriptiveness.} Finally, we model $p_{\text{reward}}(u \mid \theta)$, the second term in $P_{S_1}$, with a \emph{base speaker} model, $\speakerbase$, that maps rewards to reward descriptive utterances: $p_{\speakerbase}(u \mid \theta)$\footnote{In principle, $p_{\text{reward}}$ could also do pragmatic reasoning to optimize a listener's reward belief, but we did not find an improvement from doing so empirically.}. We also provide details on $\speakerbase$ in \secref{sec:base-models}.

\subsection{A Generative Model of Utterances}
\begin{figure}[h!]
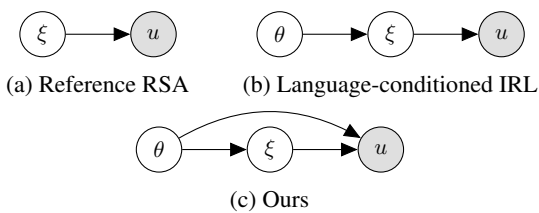

\centering
\begin{subfigure}{.4\linewidth}
    \centering
    \tikz[scale=.85, transform shape]{
     \node[latent] (x) {$\xi$}; %
     \node[obs, right=of x] (u) {$u$};%
    
     \edge {x} {u};
    }
    \caption{\centering Reference RSA}
    \label{fig:graphical-model-reference-game}
\end{subfigure}\hfill
\begin{subfigure}{.6\linewidth}
    \centering
    \tikz[scale=.85, transform shape]{
     \node[latent] (r) {$\theta$}; %
     \node[latent, right=of r] (x) {$\xi$}; %
     \node[obs, right=of x] (u) {$u$};%
     
     \edge {r} {x};
     \edge {x} {u};
    }
    \caption{Language-conditioned IRL}
    \label{fig:graphical-model-irl}
\end{subfigure}
\vspace{3mm}
\begin{subfigure}{.5\linewidth}
    \centering
    \tikz[scale=.85, transform shape]{
     \node[latent] (r) {$\theta$}; %
     \node[latent, right=of r] (x) {$\xi$}; %
     \node[obs, right=of x] (u) {$u$};%
     
     \edge {r} {x};
     \edge {x} {u};
     \draw [->] (r) to [bend left] (u);
    }
    \caption{Ours}
    \label{fig:graphical-model-ours}
\end{subfigure}
\vspace{-1em}
\caption{Graphical models contrasting prior work with our model, which models how language utterances $u$ convey both explicit information about the reward $\theta$ and implicit evidence about $\theta$ through the actions they suggest (via trajectories $\xi$). Dependence on $\cM$ not shown for visual clarity.}
\vspace{-1em}
\label{fig:graphical-models}
\end{figure}

Our account of pragmatic generation can also be viewed as the graphical model in Figure \ref{fig:graphical-models}(c), where, importantly, the reward influences the utterance both directly and via the action that the speaker refers to. We define $p(u \mid \xi, \theta, \cM)$ to be:
\begin{align*}
    p(u \mid \xi, \theta, \cM) & = \alpha p(u \mid \xi, \cM) \\ & + (1-\alpha) p(u \mid \theta, \cM)
\end{align*}
and assume that utterances are reward-descriptive in a way that is independent of the current context, $p(u \mid \theta, \cM) = p(u \mid \theta)$.

We can confirm this leads us back to $p_{S_1}$ by marginalizing out $\xi$:
\begin{align*}
    p(u & \mid \theta, \cM) = \sum_{\xi} p(u \mid \xi, \theta, \cM) p(\xi \mid \theta, \cM) \\
                         = & \alpha \sum_{\xi} \Big(p(u \mid \xi, \cM)  p(\xi \mid \theta, \cM)\Big) \\
                          & + (1 - \alpha) p(u \mid \theta, \cM) \\
                        = & \alpha p_{\text{action}}(u \mid \theta, \cM) + (1-\alpha)  p_{\text{reward}}(u \mid \theta)
\end{align*}

Using this graphical model, we illustrate how our model differs from prior work in similar settings:

\paragraph{Classic reference game pragmatics collapses belief and behavior.}
In general, RSA allows the speaker to optimize for any ``utility function,'' and in the simplest form the utility function optimizes for the listener's belief over world states \cite{GoodmanFrank2016-TICS}. However, in most work on RSA the only relevant world-state belief is belief about behavior, \eg the referent that should be selected (\figref{fig:graphical-model-reference-game}). Instead, our setting disentangles communication about intended referents in a single context and communication about (reward) beliefs, which influence behavior on longer horizons. \citet{andreas2017translating, sumers2021extending} have made the same observation: reference games conflate whether the speaker's objective is to influence beliefs or actions, and modeling the speaker as one or the other produces distinct interpretations of utterances (\eg speakers that only optimize for correct behavior may do so at the cost of being truthful about the reward).

\paragraph{IRL assumes all information about the reward function is modulated by the trajectory.}
Prior work \cite{macglashan2015grounding, fu2019language} uses IRL to recover rewards from \emph{trajectories} (\eg from datasets pairing utterances with trajectories), and then supervising a model with these induced (utterance, reward) pairs. While prior work has not specifically considered pragmatics (\ie speaker models), their implicit speaker model amounts to assuming that all information about the reward comes from trajectories, as in \figref{fig:graphical-model-irl}. In our experiments we compare against a pragmatic version of this action-centric speaker, which is equivalent to setting $\alpha=1$ in our model (only using $p_{\text{action}}$). In realistic settings where utterances are \emph{not} unambiguous commands like ``go to the red door,'' it becomes important to model how actions and utterances reveal \emph{complementary} information about rewards.

\section{The \taskname{} Task}

We design \taskname{}, a task for reward inference from natural language in the flight booking domain. \taskname{} is designed to simulate a simplified interaction with a  flight booking agent, where users communicate with the agent via language to book flights from a set of options. Effective agents must not only learn to book the preferred flight given an instruction in the immediate context (instruction following), but also learn the user's preferences over repeated interactions to book preferred flights in unseen contexts.

We collect a dataset of natural language in a multi-turn game between a user (the ``speaker'') and an assistant (the ``listener'' agent). Each flight is represented by a feature vector $\phi(\xi) \in \mathbb{R}^8$ (\eg features of carrier, price, etc.). We assume the user has a linear reward function with parameters $\theta \in \mathbb{R}^8$, specifying a reward for a particular flight $r_{\theta}(\xi) = \theta^{\intercal} \phi(\xi)$.

In the first round of the game, the user and assistant observe a set of three flight options and the user provides an utterance to describe the flight they want (the optimal flight under the reward function), \eg \utt{ ``the flight with the most stops.''} In each of the subsequent rounds, the user and assistant are presented with a new set of three flights. The assistant can either \textit{choose} by guessing the user's preferred flight (under the same reward function), or \textit{prompt} the user for another utterance describing the desired flight in the new set. If the assistant chooses but does so incorrectly, the user is prompted for another utterance describing the correct flight. Both players are penalized if the assistant chooses incorrectly, and earn points if the assistant chooses correctly (with more points for each round the assistant can do so without asking for help). The user is thus incentivized to provide utterances that inform the agent which flight to choose, while enabling long-term success over later rounds.

\subsection{Data collection}
\begin{figure}[t!]
\begin{boxedminipage}{\columnwidth}
    \begin{spacing}{0.5}
    {\small\fontfamily{cmss}\selectfont
    one stop that is short \\[0.3em]
    american is the flight that i want. but i need the flight that is the cheapest and has less stops. \\[0.3em]
    anything but american \\[0.3em]
    jetblue one \\[0.3em]
    i need a flight with any airline but jet blue, price and number of stops are a bad factor for me also. i prefer delta if affordable and low layovers. can you help me? \\[0.3em]
    even american is undesirable, paying more is important \\[0.3em]
    i like the flight that is \$64\\[-1em]
    }
    \end{spacing}
\end{boxedminipage}
    \caption{Sample text from the task, exhibiting a diversity of instructive and reward-descriptive language.}
    \vspace{-1em}
    \label{fig:sample-data}
\end{figure}
To collect data for the task, we recruit Amazon Mechanical Turk workers and randomly pair them to play six games (\ie six different reward functions) of six rounds each. Each game thus consists of 1-6 utterances describing options for the same reward function in different contexts. One person plays the role of the user and the other acts as the assistant. The user has access to a hidden reward function, which is a discretized, randomly-sampled vector $\theta \in \{-1, -0.5, 0, 0.5, 1\}^8$.
In total, we collected 2,568 utterances across 813 games, of which we split off the 91 games with the highest score (where the speaker and listener were able to communicate most effectively) for the evaluation set. More details about the data collection process can be found in \secref{sec:appendix-data-collection} of the appendix.

A sampling of text is shown in \figref{fig:sample-data}. Utterances exhibit a range of phenomena: some users lean towards describing very option-specific features (e.g. {\fontfamily{cmss}\selectfont``i like the flight that is \$64''}). Other users attempt to describe as much of their reward function as possible (e.g. {\fontfamily{cmss}\selectfont``i need a flight with any airline but jetblue,\ldots''})---we note that even when they did so, the user's tradeoffs between features remain ambiguous.
Many of the utterances are neither fully option-specific nor fully reward-descriptive: instructions like {\fontfamily{cmss}\selectfont``one stop that is short''} both instruct the agent which flight to select in the present context, while communicating some generalizable (but incomplete) information about the user's preferences.

\section{Model Implementation}
\label{sec:base-models}
Our pragmatic model (\secref{sec:model}) relies on base listener and speaker models $\listenerbase$ and $\speakerbase$.
In this section, we describe implementations of these models for the \taskname{} dataset.
To train the base models, we use the speaker-side data of (utterance, option set, reward function) tuples from each round. Our base listener and speaker models assume that the utterances are generated conditionally independently given the reward; we capture the dynamics of multiple turns in the posterior reward inference.
Both base models learn neural encodings of utterances $u$, actions $\xi$, and rewards $\theta$, and produce distributions by applying softmax functions to inner products between these encodings. 
We use $\xi^*$ to denote the optimal action in each context, \ie $\xi^* = \arg\max_{\xi} r_{\theta}(\xi)$.

\paragraph{Base listener model.}

The base listener model $\listenerbase$ is defined using inner product similarities between learned representations of actions $\xi$ produced by an MLP encoder, and learned representations of utterances produced by a BERT-base \cite{devlin-etal-2019-bert} encoder:
\[
p_{\listenerbase}(\xi \mid u, \cM) \propto \exp(\text{MLP}_{\listenerbase}(\xi) \cdot \text{BERT}_L(u))
\]
where the distribution is normalized over all actions (flights) available in the context, $\xi' \in \cM$.

We set the rationality parameter $\beta=\infty$ in $p_{\text{opt}}$ as speakers tend to refer primarily to the optimal option in our domain.

\paragraph{Base speaker model.} The base reward speaker model $\speakerbase$ is defined using an inner product between representations of rewards $\theta$ from an MLP encoder, and utterance representations from a BERT encoder:
\begin{align*}
p_{\speakerbase}(u \mid \theta) &\propto \exp (\text{MLP}_{\speakerbase}(\theta) \cdot \text{BERT}_S(u) / \tau )
\end{align*}
where $p_{\speakerbase}$ is normalized over a set of utterances taken from the training data (see \secref{sec:appendix-base-model} in the appendix), and $\tau=3$ is a temperature parameter.

\paragraph{Training.}
We fine-tune all model parameters, including the parameters of the initially-pretrained BERT utterance encoders in the listener and speaker on $(u, \xi, \cM)$ pairs from the training data using the AdamW optimizer \cite{kingma2015adam,loshchilov2017decoupled}. The listener and speaker models are trained separately, without sharing any parameters between the encoders used in the two models.
We independently train 5 random seeds of each base model and ensemble them together in evaluation by averaging their output probabilities, which we found improved performance of all models (both our full model and baselines).
See \secref{sec:appendix-base-model} in the appendix for details and model hyperparameters.

\paragraph{Pragmatic inference} We follow previous work \citep{fried2017unified, monroe2017colors} and approximate the $S_1$ distribution by normalizing over a fixed set of utterances: the de-duplicated set of short utterances (less than 8 tokens, making up the majority of utterances) with no digits from the training data. We implement the full pragmatic model $p_{L_2}(\theta \mid u, \cM)$ in Pyro \citep{bingham2018pyro} and use importance sampling to generate samples from the posterior over rewards. Given our dataset collection procedure (where we uniformly sample rewards), we model an uniform prior over rewards $p(\theta \mid \cM)$ for the first interaction.
\section{Experiments}
\begin{figure*}[ht]
\centering
\begin{subfigure}{.48\textwidth}
  \centering
  \includegraphics[width=\linewidth]{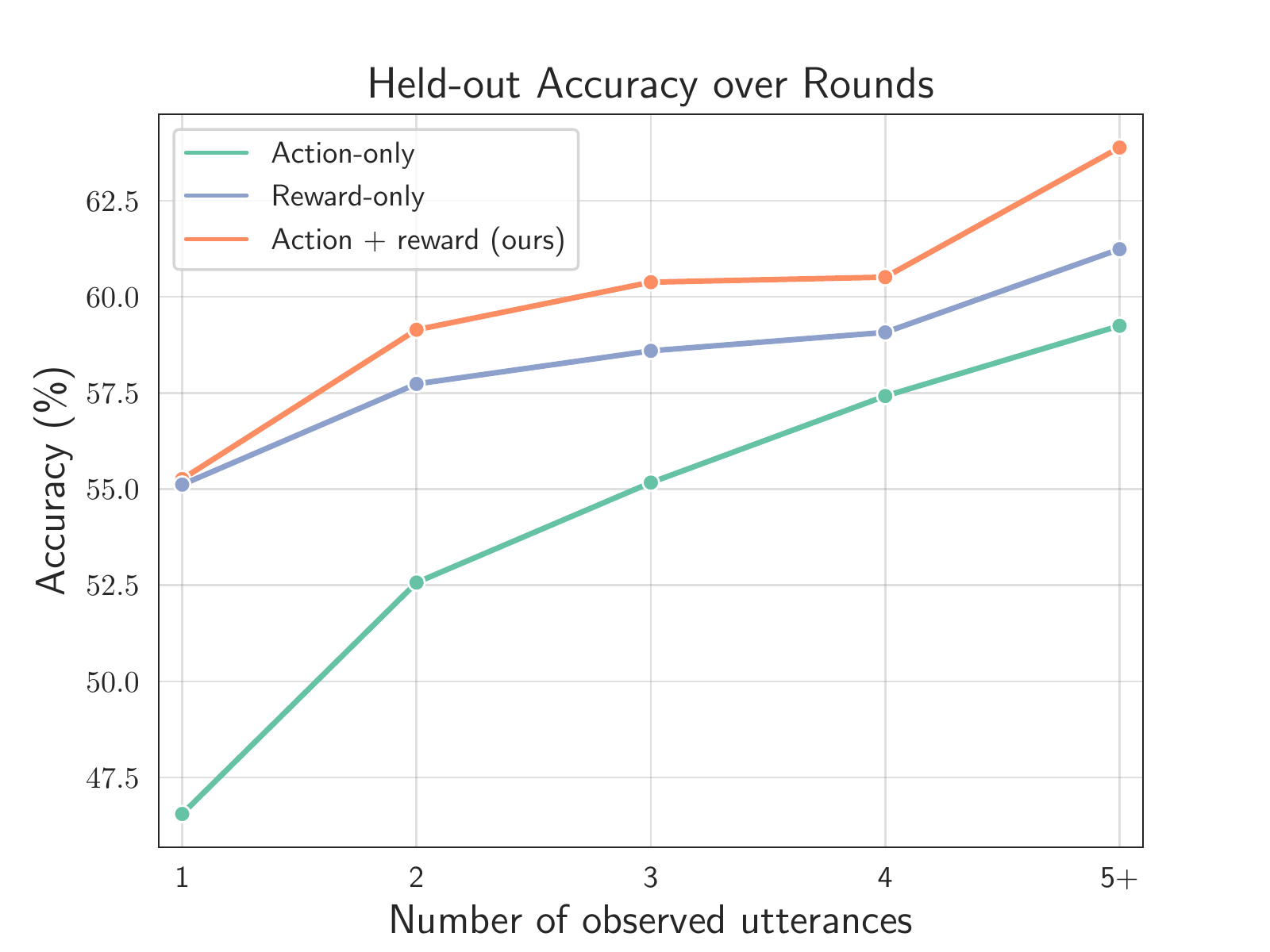}
  \label{fig:multiturn-acc}
\end{subfigure}%
\begin{subfigure}{.48\textwidth}
  \centering
  \includegraphics[width=\linewidth]{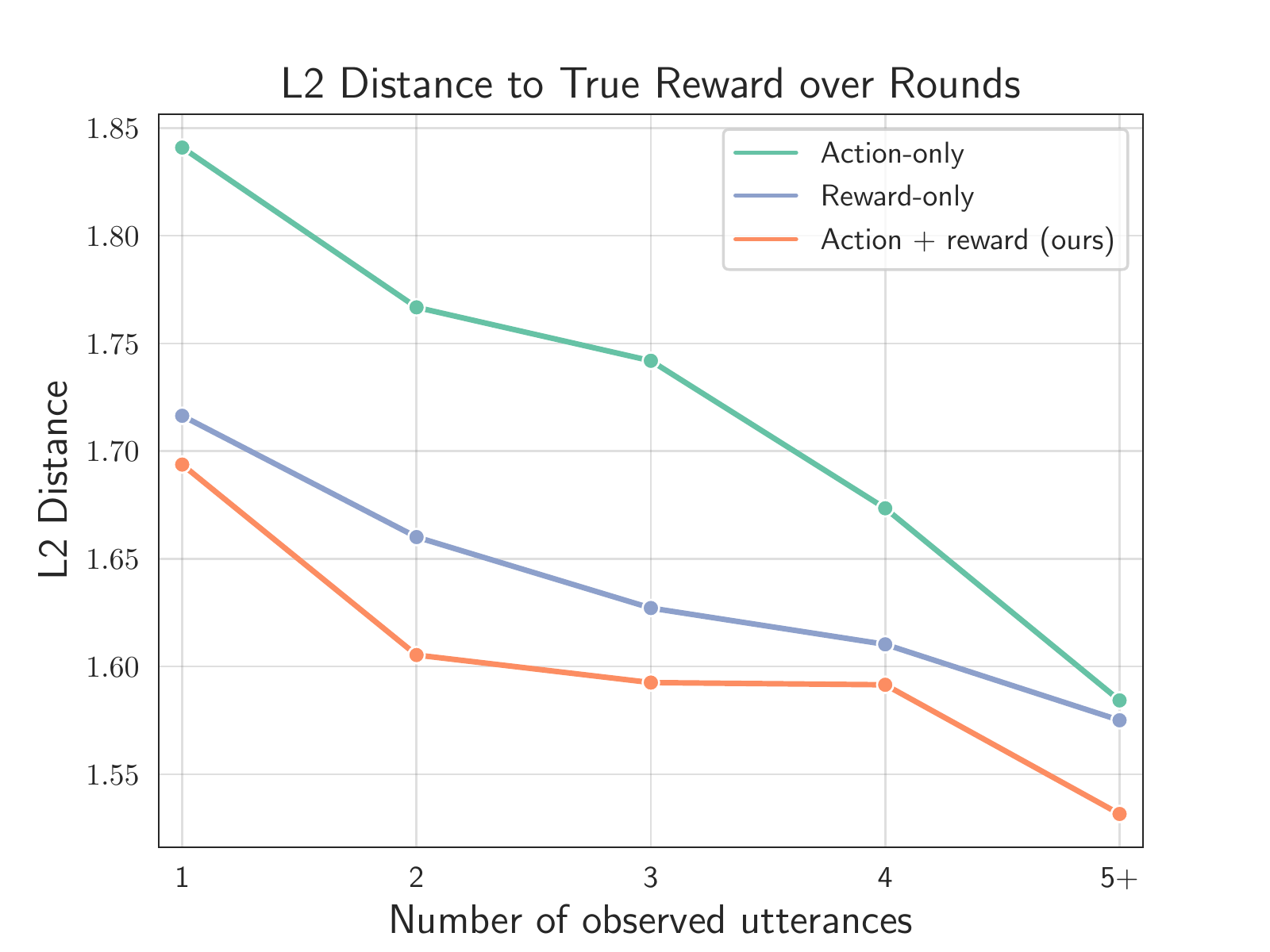}
  \label{fig:multiturn-l2}
\end{subfigure}
\vspace{-1em}
\caption{\textbf{Multi-turn performance on held-out accuracy (left) and L2 distance to the true reward (right).} We show the performance of each model for varying numbers of observed utterances for a given reward. We combine five- and six-utterance rounds as there were $<25$ samples in each of these bins. Our full action+belief model substantially outperforms an action-only model at all numbers of utterances ($p < .05$), and performs comparably to or better than a belief-only model, with statistically significant benefits for 5+ utterances ($p < .05$).
}
\vspace*{-1em}
\label{fig:multiturn}
\end{figure*}
\begin{table}[t]
\centering
\resizebox{\columnwidth}{!}{
    \begin{tabular}{lc}
    \toprule
    \textbf{Method} & \textbf{Held-out accuracy (\%)} \\
    \midrule
    \multicolumn{2}{l}{\textit{Oracle models} (infer $k$ features perfectly)} \\
    \hspace{3mm}$k=1$                   & 43.0 \\
    \hspace{3mm}$k=2$                   & 51.5 \\
    \hspace{3mm}$k=3$                   & 60.2 \\
    \hspace{3mm}$k=4$                   & 64.7 \\
    \midrule
    Action-only             & 52.8 $\pm$ 0.97 \\
    Reward-only             & 57.8 $\pm$ 0.95 \\
    Action + reward (Ours)  & 59.1 $\pm$ 0.96 \\
    \bottomrule
    \end{tabular}
}
\caption{Average held-out accuracy averaged over all evaluation rounds, with standard error of the mean indicated. Our full action+reward model significantly outperforms action-only and reward-only models (with $p<.05$ using the paired bootstrap test). Held-out accuracy is also shown for oracle models that infer $k$ (randomly-chosen) features of the reward perfectly and maintain a uniform distribution over the other features.}
\vspace{-1em}
\label{tab:overall-acc}
\end{table}

We evaluate models in the same repeated turn setup that humans carried out in the task. For each game, models play the role of the listener in that game, updating the reward posterior (\secref{sec:model}) after observing the utterance and option set in each round. 
Our goal is to estimate rewards that allow the agent to carry out the person's preferences: choosing the optimal option (flight) in unseen contexts (sets of flight options). To that end, we directly compare models on \textbf{held-out accuracy}: on 1,000 randomly-generated sets of three options, how often the model's estimate of the reward, $\hat \theta$, selects the option that is optimal under the true reward.\footnote{Note that when collecting the dataset, we also tested human listeners's ability to generalize, but only had them select an option on a single unseen option set---the next one in the sequence---to make data collection tractable.}
We use the model's reward posterior mean as the estimate, $\hat \theta = \mathbb{E}_{p_{\theta}} \theta$. We additionally provide comparisons of \textbf{reward L2 distance} between the estimated reward and the true reward as a context-independent metric:
$\sqrt{\sum_{i=1}^8 (\hat{\theta_i} - \theta_i^{*})^2},$ where $\theta^{*}$ is the true reward.

For our full \emph{action + reward} model, we set the nearsightedness parameter $\alpha=0.5$ for all posterior updates. 
We compare to an \textit{action-only} model that uses only $p_{\text{action}}$ (\ie setting $\alpha=1.0$). This model is representative of approaches from past work on language-conditioned reward learning (\eg \citealt{macglashan2015grounding,fu2019language}) that infer rewards purely from the actions that utterances refer to. We also compare to a \textit{reward-only} model that uses only $p_{\text{reward}}$ (inferring rewards purely from the utterance, without conditioning on actions, \ie setting $\alpha=0.0$).
For comparison to versions of our approach that remove pragmatic modeling, see \secref{sec:appendix-pragmatic-analysis} in the appendix.

\subsection{Overall Results}
In \tabref{tab:overall-acc} we compare all models on held-out accuracy averaged over all rounds in the evaluation set (for each round, having observed all previous rounds in that game). Note that because held-out accuracy is assessed by the proportion of randomly-generated flight sets (out of 1,000) where the true reward function and the inferred reward function pick out the same optimal flight, it is significantly more difficult than achieving high accuracy on a single three-choice instance.

Our full action+reward model achieves a held-out accuracy of 59.1\%, +6.3\% over the action-only model and +1.3\% over the reward-only model, indicating that combining both sources of information allows better inference of rewards that enable optimal actions in novel contexts. For reference, an oracle baseline that infers the value of $k$ randomly chosen features perfectly and is uniform on the other features obtains the following held-out accuracies: $k=$\textbf{1} (43\%), \textbf{2} (51\%), \textbf{3} (60\%), \textbf{4} (65\%), showing that our model is able to attain similar generalization performance even in the presence of uncertainty (without receiving oracle information about the true value of any feature).

We analyze why our model benefits from both components in \secref{sec:exp-benefits}, and discuss potential for further improvements in \secref{sec:improving-models}.

\begin{figure*}[ht]
\centering
\begin{subfigure}{.45\textwidth}
  \centering
  \includegraphics[width=\linewidth]{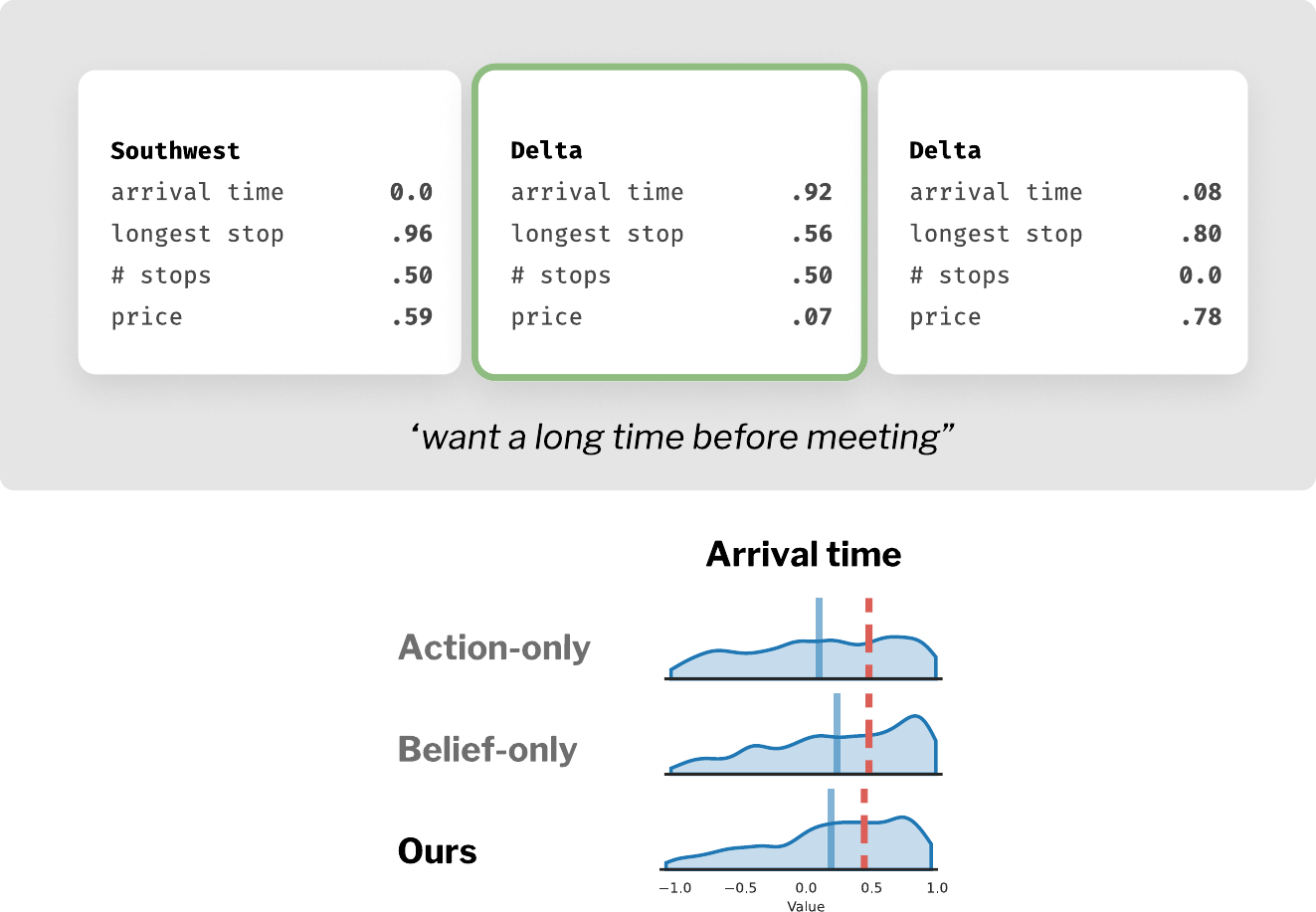}
  \caption{Both the described action (the referenced flight is the one with the highest arrival time) and the explicit reward description in the utterance provide evidence that the user's true reward on arrival time is positive, leading the posterior in our model to (correctly) place more probability mass on positive values of this feature.}
  \label{fig:prag-example-1}
\end{subfigure}\hfill%
\begin{subfigure}{.45\textwidth}
  \centering
  \includegraphics[width=\linewidth]{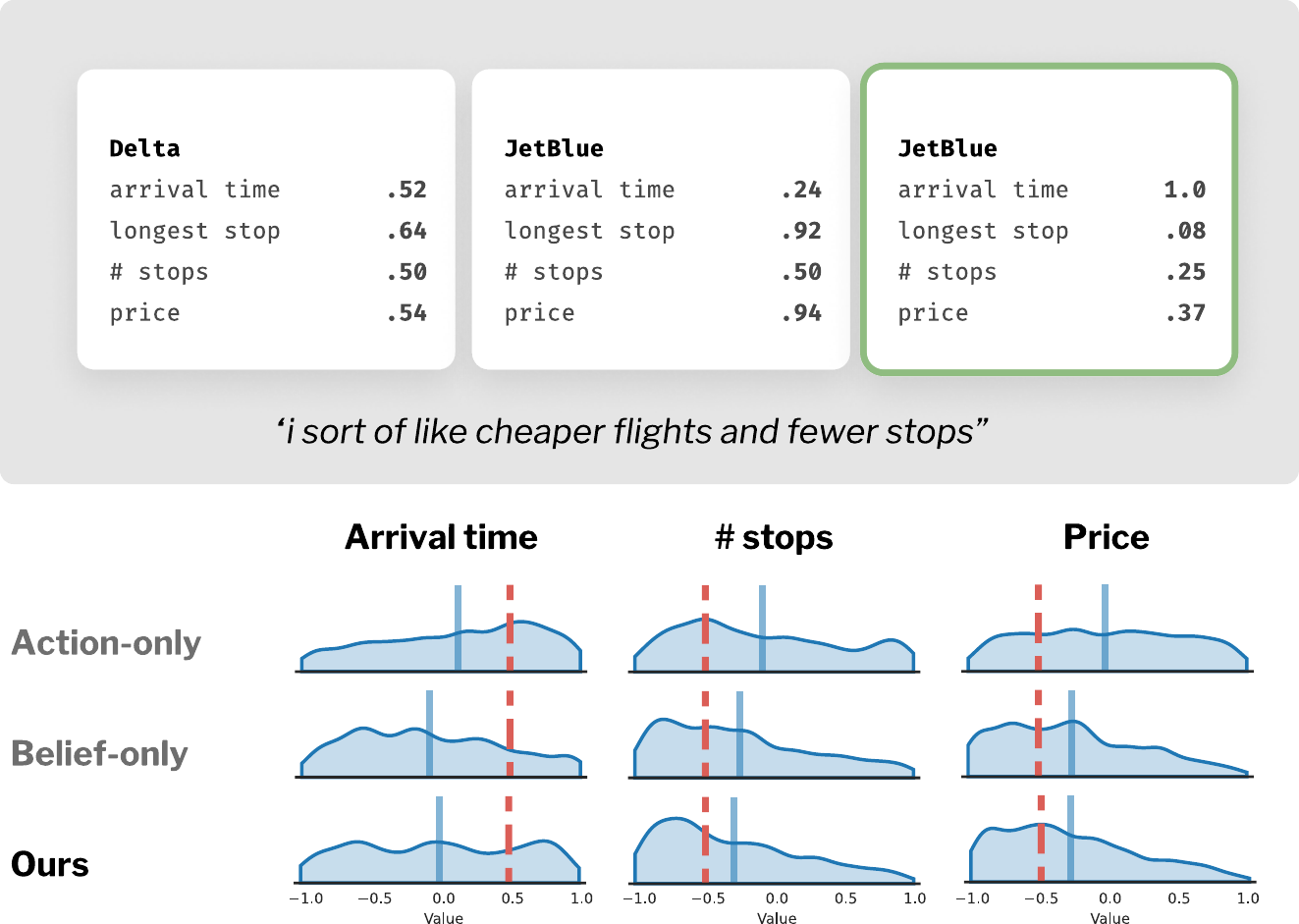}
  \caption{Evidence from actions and from the utterance complement each other: the action-based model captures that rewards that are positive on arrival time make the selected flight optimal, even though it is unmentioned, while the reward-based model captures evidence about the reward from the user's utterance.}
  \label{fig:prag-example-2}
\end{subfigure}
\caption{\textbf{Real examples showing reward posteriors of each model after observing the given utterance and options.} We sample from the posterior over rewards and visualize the marginal probability distributions for particular features using kernel density estimation. The true reward value for the feature is marked with a {\color{red}{red}} line and the posterior mean for the feature with a {\color{blue}{blue}} line.}
\label{fig:pragmatic-examples}
\vspace{-1em}
\end{figure*}

\subsection{Learning over Multiple Interactions}
\label{}

We explore how each model's reward inferences change as more observations are obtained over the course of a game. In \figref{fig:multiturn}, we plot held-out accuracy and L2 distance to the true reward as a function of number of observed utterances.
Our model outperforms the action-only and reward-only models for all numbers of observed utterances.

\paragraph{Relying on explicit information from language is most important when there are few observations.}
While our full action+reward model improves substantially over the action-only model at all points, this improvement generally decreases as more utterances are observed (\figref{fig:multiturn}). Conversely, the improvement of the full model over reward-only generally increases.
Qualitatively, we observe that this occurs because utterances tend to mention the most extreme features of the reward function, which allow our model to estimate the values of these important features. When there are few observations, inferring reward information from utterances in this way is more informative than using only the option implied by the user's utterance, which does not disambiguate between rewards that select the same option (a commonly discussed problem in IRL; \citet{ziebart2008maximum}).

\paragraph{Inferring evidence from actions is most important when there are more observations.} We observe that the action-only model improves more consistently over rounds. Qualitatively, the information that utterances provides about rewards is correlated across multiple rounds---speakers frequently mention salient reward features, 
whereas actions consistently provide new information about all features. This is particularly pronounced in our domain, due to a relatively small feature and action space. In other more complex domains, actions might provide even more benefits as they provide \emph{fine-grained} information about reward values and tradeoff boundaries that are more difficult to communicate precisely in language.

\subsection{Analyzing the Benefits of Combining Actions and Rewards}
\label{sec:exp-benefits}

In this section, we investigate \emph{why} our model benefits from both the action and reward models.

\paragraph{A single utterance and context can provide useful evidence to both models.} In \figref{fig:pragmatic-examples}, we show the reward posteriors for each model after a single update on a round (starting from a uniform prior). In \figref{fig:prag-example-1}, we observe how the action- and reward-only models can make correlated updates on an utterance and context where both the action (a flight with a high value on arrival time) and the utterance provide evidence about the arrival time feature. This leads our model's posteriors to aggregate more probability mass on positive values of that feature. In \figref{fig:prag-example-2}, we show how each model can make inferences about different features for the same context---the action-only model inferring positive values for arrival time given the observed flight and the reward-only model updating on flight price and stops. Our model posterior aggregates information from both.

\paragraph{Some utterances are primarily ``nearsighted,'' and others primarily ``farsighted.''}
Another reason our full model improves is because some utterances are particularly ``farsighted''---mentioning a great deal of explicit information about the reward (which the action-only model cannot take advantage of)---while other utterances are more ``nearsighted''---specialized to the particular action, \eg saying just enough to uniquely identify the optimal flight.
Sorting the utterances by difference in accuracy between the action-only and reward-only models confirms that they exhibit qualitatively different phenomena: examples where the reward-only model helps the most are highly reward-descriptive (\eg \utt{ ``if i had a choice, i would never fly with delta and american! get me jetblue or southwest\ldots}'') while examples where the action-only model helps most have less informative utterances (\eg \utt {``the cheaper the better''}).
Our full model is able to handle both kinds of language use.

To further analyze the influence of the action and reward component, we evaluate an oracle model that \textit{switches} between the action-only and reward-only models, choosing the model with highest held-out accuracy in each round. This model outperforms our action+reward model (improving from 59.1 to 62.9\% on overall held-out accuracy), suggesting that further improvements could be obtained by integrating evidence from the two models. 
Doing so optimally is challenging in our setting: when a user says ``\utt{i like the cheap jetblue flight},'' do they mean to say they like JetBlue generally, or just that they want to choose a desirable flight that happens to be uniquely identified by JetBlue?  Future work might explore adaptively switching policies (\eg using the utterance, or knowledge about the user).

\subsection{Inference Improves with Known Actions}
\label{sec:improving-models}
While our base models have fairly high performance (\eg the base listener model $\listenerbase$ has an average accuracy of 74\% at selecting the optimal choice in each option set that has an utterance in the evaluation data), they naturally have some errors which lead to errors in reward inference.
We test the influence of this underlying prediction error by skipping posterior updates on all rounds where the base listener predicts the incorrect option for the true reward function.
This change improves held-out accuracy by 6\% over the reward-only model after six observations (+4\% from the original gap), indicating (1) that dataset affords future work on improved instruction following models and (2) that our reward inference procedure benefits from base model improvements.

We note that in our task design, the user does not provide a demonstration (\ie a choice of flight) to the model. However, if it is convenient to obtain demonstrations from users (\eg a flight booking interface could let the person click on the flight they want in addition to specifying what they want in natural language), demonstrations would effectively serve as an oracle instruction-following model for that context, which could be incorporated into our full reward inference model.

\section{Discussion \& Conclusion}

We presented a method for using natural language to infer reward functions: representing the goals, preferences, and intents underlying action. 

Conceptually, our work builds on previous work on language grounding by exploring how language serves a dual purpose. Utterances can refer directly to actions to be taken, as studied in instruction following. Beyond that, they communicate information about ``why'' those actions should be taken, and what actions may be desirable in new contexts. To build language-guided agents that can interact with people over longer horizons, it may be useful to model this relationship between language, actions, and rewards.

Furthermore, language is \emph{ambiguous} about both actions and goals. Standard settings for studying pragmatics (\eg reference games) address how to resolve ambiguity about what object or action the speaker is choosing to refer to. We have explored how these settings can be extended by considering the preferences underlying those choices. We introduced \taskname{}, a new dataset of naturalistic interactions between people in a multi-turn flight booking game. \taskname{} uses held-out accuracy as a metric for evaluating interpretation success beyond selecting the right action in a single environment.

Future work can build on the task by 1) learning or evaluating with more complex reward functions (\eg using deep reward representations); 2) exploring how people communicate about their real preferences and modeling a natural prior (\eg that people tend to prefer cheaper flights), instead of providing annotators with ground-truth preferences; 3) allowing other ways to handle uncertainty, \eg leveraging the reward posterior to interactively learn to ask; or 4) extending these approaches to other domains where modeling goals and preferences may be important (\eg language-conditioned robotics).

\section*{Acknowledgements}
We thank Eric Wallace, Jerry He, and the other members of the Berkeley NLP group and InterACT Lab for helpful feedback and discussion. This work is supported by a grant from the Office of Naval Research (ONR-YIP).

\bibliography{anthology,custom}
\bibliographystyle{acl_natbib}

\clearpage
\appendix

\section{Data Collection}
\label{sec:appendix-data-collection}

 We recruited Amazon Mechanical Turk workers from the US with an approval rate of $\geq$ 98\%, $\geq$ 5,000 Human Intelligence Tasks (HITs) completed, and completion of a custom qualification task where they played 15 minutes of the game and demonstrated active participation from manual review. Turk workers were given the following instructions for the task (shortened):

\textit{Scenario: A customer working with a new personal assistant to book flights for their business meetings (you will play either the customer or the assistant, and another Turker will play the other role).}

\textit{As the customer, you have specific preferences for what kind of flights you like and dislike. When you first start working with your assistant, you might need to tell them exactly what you want, but you hope that over time, they will figure out what you like and book your preferred flights without your help (imagine an assistant that knows you well enough to say: “Bob hates redeyes and doesn’t like to be rushed on the way to the airport, so I’ll go ahead and book this 2pm Jetblue to New York.”)}

\textit{As the assistant, you want to figure out what the customer likes and book flights that they want. Pay attention to what they say when they choose flights.}

\textit{What is this task for? The goal of this task is to study how people naturally communicate their preferences to personal assistants, with the goal of building digital assistants (like Siri) that better understand and learn what people want. For the purposes of the task, we will give the customer “fake” flight preferences. If you are the customer, pretend that these are the kinds of flights you actually like / dislike.}

The interface for the task (for an assistant) is shown below:

\begin{figure}[h]
\centering
\includegraphics[width=.9\linewidth]{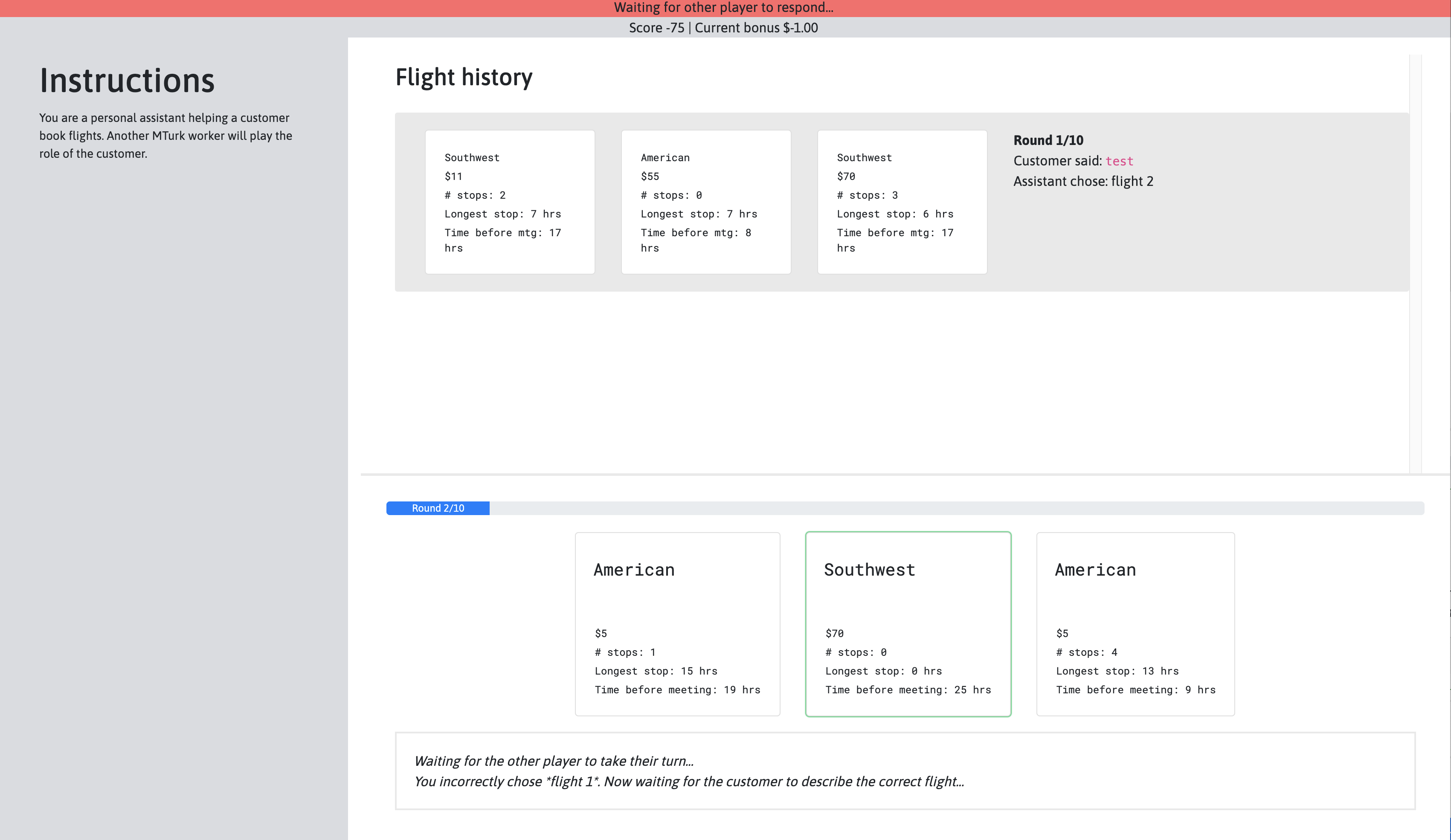}
\end{figure}

Workers were compensated \$4 for 30 minutes (6 games of 6 rounds each), with a \$0.20 bonus for every 25 points. For each round, points were accumulated based on the assistant's action:
\begin{itemize}
    \item Assistant chooses correctly: +25 points
    \item Assistant chooses incorrectly: -100 points
    \item Assistant asks for help: -20 points
\end{itemize}

\section{Dataset Details}

We plot the distribution of utterance lengths (number of tokens) in our dataset below:
\begin{figure}[h]
\centering
\includegraphics[width=.9\linewidth]{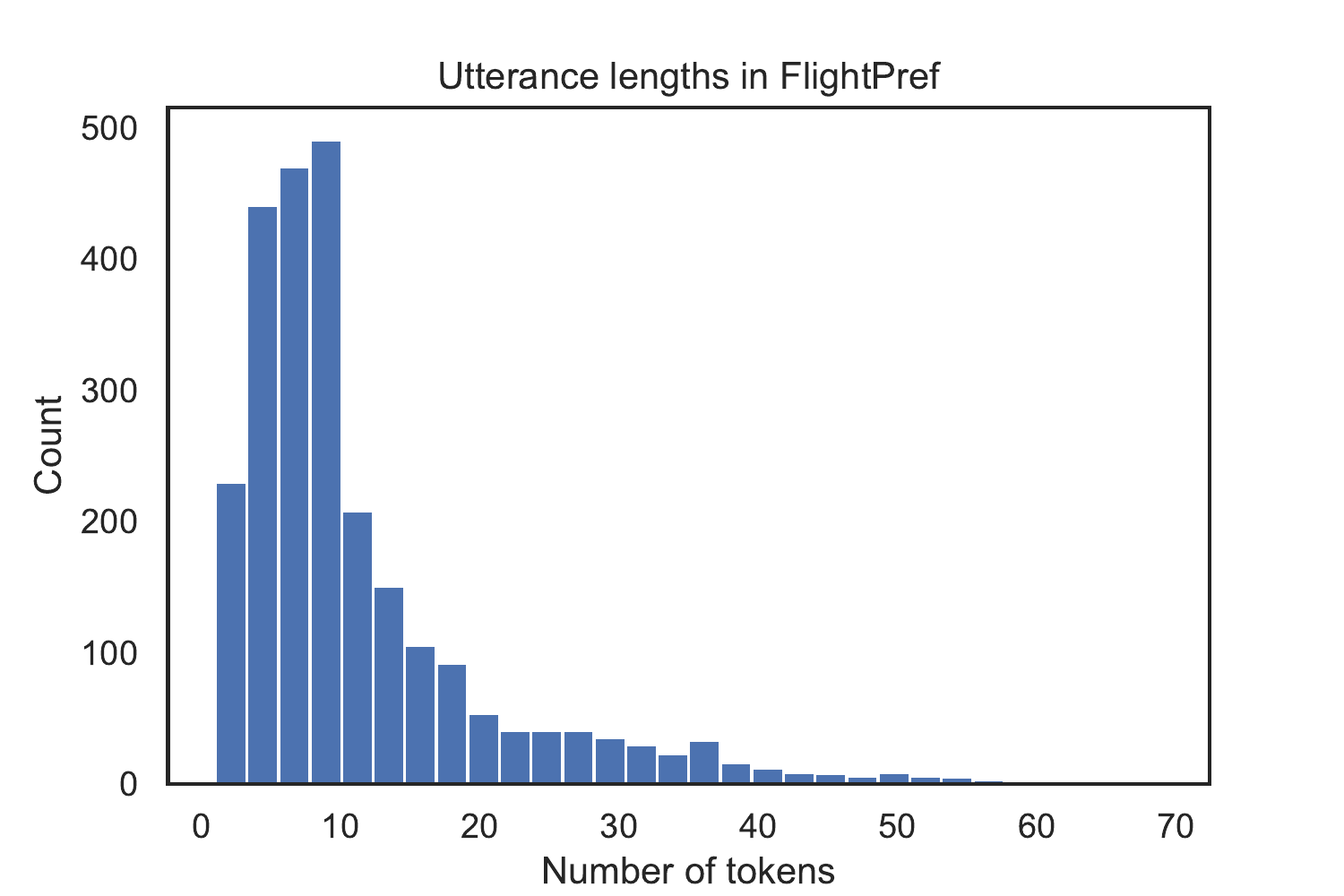}
\end{figure}

\section{Base Model Details}
\label{sec:appendix-base-model}

\subsection{Training}

We fine-tune all model parameters, including the parameters of the initially-pretrained BERT utterance encoders in the listener model and speaker model.
We produce utterance representations from BERT using a linear projection of BERT's \texttt{[CLS]} embedding for the utterance.
Models are trained separately on the training data using the loss functions below, with an AdamW learning rate of $2 \times 10^{-5}$ and a batch size of 64 for the listener model and a learning rate of $5\times 10^{-5}$ and a batch size of 32 for the speaker model. We perform early stopping using the loss on held-out validation data.

\paragraph{Listener loss.}
We define the following loss function for the base listener model $\listenerbase$ on each training instance:
\[
\mathcal{L}_{\listenerbase} = - \log p_{\listenerbase}(\xi \mid u, \cM)
\]
\paragraph{Speaker loss.}
It would be possible to learn the parameters of the base reward speaker $\speakerbase{}$ directly on the training data, using a loss function similar to the base listener model. However, since utterances are often also action-descriptive, utterances cannot typically be predicted accurately from rewards alone.
To account for this, we also define a separate base \emph{action speaker} model, $p_{S_{\text{act}}}(u \mid \xi^*, \cM)$ that produces utterances conditioned on the optimal action $\xi^*$ in context $\cM$: 
\[
p_{S_{\text{act}}}(u \mid \xi^*, \cM) \propto \exp (\text{MLP}_{S_{\text{act}}}(\xi^*) \cdot \text{BERT}_S(u)) 
\]
where $p_{S_{\text{act}}}$ is normalized over the same set of utterances as $p_{\speakerbase}$. The base action speaker model is used only in training. It would be possible to also use this base action speaker in evaluation, in place of the pragmatic action speaker $p_{\text{action}}$ (\secref{sec:model}); however, we found that pragmatic reasoning about a conventional instruction following model, as outlined in \secref{sec:model}, performs better.

We train the two base speaker models jointly using a simple latent variable model, which makes the simplifying assumption that every utterance is either action- or reward-descriptive. To model this, we use a discrete latent variable $\lambda \in \{0, 1\}$:
\begin{align*}
p(u \mid \theta, \xi^*, \cM, \lambda) =& \lambda p_{\speakerbase}(u \mid \theta) + \\
&(1 - \lambda) p_{S_{\text{act}}}(u \mid \xi^*, \cM)
\end{align*}
The loss function for a single example is
\[
\mathcal{L}_{\speakerbase} = -\log \sum_{\lambda \in \{0, 1\}} p(\lambda) p(u \mid \theta, \xi^*, \cM, \lambda)
\]
where $p(\lambda)$ gives the probability of an utterance being reward-descriptive or action-descriptive. We model this using $p(\lambda=1) = \sigma(l)$, where $l$ is a parameter which is updated in training.
Intuitively, the latent variable model learns soft clusters for action-descriptive and reward-descriptive utterances, with reward-descriptive utterances providing stronger supervision for the $\speakerbase$ model.

\paragraph{Speaker normalization.}
In training, we compute the normalizers for the $\speakerbase$ model using all utterances within the mini-batch, as well as up to 4 synthetically-constructed hard-negative examples defined by replacing attribute mentions  within the true utterance (detected using exact word match) with alternative distractor attributes (\eg replacing any occurrences of ``jetblue'' with one of ``southwest'', ``american'', or ``delta'', randomly sampled). We found that constructing hard-negatives in this way allowed us to train the base speaker models effectively despite using a fairly small dataset and small training batch sizes.

In evaluation, we compute normalizers for the $\speakerbase$ model using a filtered set of all utterances from the training data that contain no more than 8 words and no digits. (We use a smaller set of normalizers in training time for efficiency reasons.)

\subsection{Hyperparameters}
\label{sec:appendix-hyperparameters}
MLPs use fully-connected layers with \texttt{ReLU} non-linearities, and dropout applied to each hidden representation during training. We show hyperparameters for the models in Table~\ref{tab:hyperparameters}.  The BERT model is BERT-base, implemented in HuggingFace's Transformers library \cite{wolf2020transformers}.
\begin{table}[h]
  \small
  \centering
  \begin{tabular}{lc}
    \toprule
    \multicolumn{2}{l}{\bf Listener Hyperparameters} \\
    $\text{MLP}_{\listenerbase}$ hidden layers & 2 \\
    $\text{MLP}_{\listenerbase}$ hidden size & 768 \\
    $\text{MLP}_{\listenerbase}$ output size & 768 \\
    $\text{MLP}_{\listenerbase}$ dropout & 0.1 \\
    \midrule
    \multicolumn{2}{l}{\bf Speaker Hyperparameters} \\
    $\text{MLP}_{\speakerbase}$ hidden layers & 2 \\
    $\text{MLP}_{\speakerbase}$ hidden size & 512 \\
    $\text{MLP}_{\speakerbase}$ output size & 128 \\
    $\text{MLP}_{\speakerbase}$ dropout & 0.2 \\
    \bottomrule
\end{tabular}
\caption{Hyperparameters for the base speaker and listener models.}
\label{tab:hyperparameters}
\vspace*{-1em}
\end{table}

\section{Analysis}
\label{sec:appendix-analysis}
\subsection{Analyzing the effect of pragmatically modeling the speaker}
\label{sec:appendix-pragmatic-analysis}
\begin{table}[b]
\centering
\small
\resizebox{\columnwidth}{!}{
    \begin{tabular}{lc}
    \toprule
    \textbf{Method} & \textbf{Held-out accuracy (\%)} \\
    \midrule
    Action + reward (Ours)  & 59.1 $\pm$ 0.96 \\
    $p(\theta \mid u)$ inference  & 53.6 $\pm$ 1.10 \\
    $p(\theta \mid u)$ training  & 52.7 $\pm$ 0.92 \\
    \bottomrule
    \end{tabular}
}
\caption{Average held-out accuracy (over 1000 rounds) of our model compared to ablated baselines that do not explicitly calculate a normalized utterance distribution $p(u\mid\theta)$, averaged over all validation rounds (each with varying numbers of observations). Standard error of the mean indicated and $p<.05$ for all observed differences using the paired bootstrap test.}
\label{tab:utt-pragmatics}
\end{table}
While we do not expect pragmatically modeling an action-only speaker will help in our domain since the action space is small (there is little ambiguity in what the referenced action is), we explore the effect of pragmatically modeling a belief-only speaker. We compare the belief-only model to two non-pragmatic alternatives that directly infer $p(\theta \mid u)$ without explicitly calculating the speaker's distribution over utterances $p(u \mid \theta)$: (1) inference: normalizing the logits of the $\speakerbase$ model over rewards $\theta$ rather than utterances, and (2) training: a $\speakerbase$ model trained to maximize $p(\theta \mid u)$ instead of $p(u \mid \theta)$. We show the results in \tabref{tab:utt-pragmatics}: our model outperforms on held-out accuracy by 5-6\% over non-pragmatic alternatives, suggesting that modeling the speaker distribution is helpful for interpreting utterances more accurately.

\subsection{Examples ranked by action-only and belief-only accuracy difference}
\begin{figure}[t!]
\begin{boxedminipage}{\columnwidth}
    {\small
    \textit{Action-only better} \\
    
    +0.37: \utt{the cheaper the better} \\
    +0.27: \utt{like american airlines best} \\
    ...\\
    \utt{american} \\
    \utt{cheapest one please} \\
    ...\\
    -0.45: \utt{i love american and like southwest. i don't like jetblue. i like low number of stops, but i like long stop times.} \\
    -0.52: \utt{if i had a choice, i would never fly with delta and american! get me jetblue or southwest if possible! if i didn't have a choice, i really like having long stopovers, so i can rest or sightsee. i also like having some time before meetings so i'm not rushed.} \\
    
    \textit{Belief-only better}
    }
\end{boxedminipage}
    \caption{Utterances with the largest difference between action-only and belief-only held-out accuracy after updating independently on that round (compared to two random other utterances in the validation set), for examples where they both do better than chance (.33). The difference $\texttt{(action\_acc - belief\_acc)}$ is shown to the left of each example.
    }
    \label{fig:near-far-diffs}
\end{figure}
Figure \ref{fig:near-far-diffs} shows utterances with the largest difference between action-only and belief-only held-out accuracy after updating independently on that round (compared to two random other utterances in the validation set), for examples where they both do better than chance (.33). The belief-only model excels at reward-descriptive utterances, whereas action-only flight tends to outperform when there is less information in the utterance.

\end{document}